\documentclass{article} 
\usepackage{misc/iclr2020_conference}
\iclrfinalcopy 

\usepackage{lipsum}
\usepackage{xcolor}
\definecolor{darkblue}{rgb}{0.0, 0.0, 0.55} 
\definecolor{darkcerulean}{rgb}{0.03, 0.27, 0.49}
\definecolor{darkcandyapplered}{rgb}{0.64, 0.0, 0.0}
\definecolor{darklavender}{rgb}{0.45, 0.31, 0.59}
\definecolor{darkmagenta}{rgb}{0.55, 0.0, 0.55}

\usepackage{overpic}
\usepackage{rotating}
\usepackage{cancel}
\usepackage{subfigure}

\usepackage[utf8]{inputenc}
\usepackage[english]{babel}
\usepackage[T1]{fontenc}
\usepackage{microtype}      
\usepackage{times}

\usepackage{graphicx,adjustbox,subcaption,wrapfig}
\usepackage{multirow,multicol,booktabs}
\usepackage{url}
\usepackage[colorlinks=true,
            linkcolor=darkblue,
            citecolor=darkblue,
            filecolor=magenta,
            urlcolor= darkmagenta,]{hyperref}
\usepackage{bookmark} 

\usepackage{natbib}
\setcitestyle{authoryear,open={[},close={]}}

\usepackage{algorithm}
\usepackage{algpseudocode}

\usepackage{listings}
\definecolor{codegreen}{rgb}{0,0.6,0}
\definecolor{codegray}{rgb}{0.5,0.5,0.5}
\definecolor{codepurple}{rgb}{0.58,0,0.82}
\definecolor{backcolour}{rgb}{0.95,0.95,0.92}
\lstdefinestyle{mystyle}{
    backgroundcolor=\color{backcolour},   
    commentstyle=\color{codegreen},
    keywordstyle=\color{magenta},
    numberstyle=\tiny\color{codegray},
    stringstyle=\color{codepurple},
    basicstyle=\ttfamily\footnotesize,
    breakatwhitespace=false,         
    breaklines=true,                 
    captionpos=b,                    
    keepspaces=true,                 
    showspaces=false,                
    showstringspaces=false,
    showtabs=false,                  
    tabsize=2
    }
\usepackage{amssymb,amsthm}
\usepackage{amsmath,amsfonts, mathtools, bm, bbm, nicefrac}
\usepackage[thinc]{esdiff}

\usepackage[capitalize,noabbrev]{cleveref}

\DeclareMathAlphabet{\mathsfit}{\encodingdefault}{\sfdefault}{m}{sl}
\SetMathAlphabet{\mathsfit}{bold}{\encodingdefault}{\sfdefault}{bx}{n}

\newcommand{\method}{AVDM2}
\newcommand{\DMD}{SDM}

\title{Accelerating Video Diffusion Models via Distribution Matching}

\author{
Yuanzhi Zhu\textsuperscript{1}\thanks{Work done while interned at Rhymes.AI (zyzeroer@gmail.com)}, 
~Hanshu Yan\textsuperscript{1}\thanks{Project leader (hanshu.yan@outlook.com)}
~Huan Yang\textsuperscript{1}, 
~Kai Zhang\textsuperscript{2}, 
~Junnan Li\textsuperscript{1}\thanks{Corresponding author}, \\
\textsuperscript{\textbf{1}}Rhymes.AI, 
~\textsuperscript{\textbf{2}}Nanjing University\\
}

\begin{document}
\maketitle


\newcommand{\yz}[1]{\textcolor{blue}{\scriptsize [YZ: #1]}}
\newcommand{\hs}[1]{\textcolor{red}{\scriptsize [Hanshu: #1]}}


\begin{abstract}
Generative models, particularly diffusion models, have made significant success in data synthesis across various modalities, including images, videos, and 3D assets. However, current diffusion models are computationally intensive, often requiring numerous sampling steps that limit their practical application, especially in video generation. This work introduces a novel framework for diffusion distillation and distribution matching that dramatically reduces the number of inference steps while maintaining—and potentially improving—generation quality.
Our approach focuses on distilling pre-trained diffusion models into a more efficient few-step generator, specifically targeting video generation. By leveraging a combination of video GAN loss and a novel 2D score distribution matching loss, we demonstrate the potential to generate high-quality video frames with substantially fewer sampling steps. 
To be specific, the proposed method incorporates a denoising GAN discriminator to distil from the real data and a pre-trained image diffusion model to enhance the frame quality and the prompt-following capabilities.
Experimental results using AnimateDiff as the teacher model showcase the method's effectiveness, achieving superior performance in just four sampling steps compared to existing techniques. 

\end{abstract}    

\section{Introduction}

Generative models have witnessed remarkable progress in recent years, with diffusion models emerging as a groundbreaking approach to data synthesis across diverse modalities \cite{sohl2015deep,ho2020denoising,nichol2021improved,dhariwal2021diffusion,song2020score}.
From photorealistic image generation \cite{ramesh2022hierarchical,StabilityAI2021,betker2023improving,podell2023sdxl,Flux2024} to video creation \cite{guo2023animatediff,zhou2024allegro,yang2024cogvideox,OpenSoraPlan2024}, and extending to emerging 3D asset generation \cite{hong2023lrm, xu2024instantmesh, wang2024llama} and beyond \cite{kong2020diffwave,yim2023se,abramson2024accurate,chi2023diffusion}, these models have demonstrated an unprecedented ability to capture complex data distributions.

Unlike Generative Adversarial Networks (GANs) \cite{goodfellow2020generative,karras2019style}, Normalizing Flows (NFs) \cite{dinh2016density} and Variational Autoencoders (VAEs) \cite{kingma2013auto,razavi2019generating} which generate new data directly by mapping from noise to data, diffusion models divide the mapping into multiple steps, enabling stabling training and better distribution matching.
However, this nature also requires multiple sampling steps for diffusion model, which limit their practical usage especially for video generative models, due to the data size and model size. 
It may take more than 10 minutes to generate a high-resolution video using 50 diffusion sampling steps with the current most advanced graphic card \cite{OpenSoraPlan2024,zhou2024allegro}.

Viewing diffusion sampling as solving the Probability Flow Ordinary Differential Equation (PF-ODE) has inspired advanced samplers like DDIM~\cite{song2020denoising}, DPM-solver~\cite{lu2022dpm}, and DEIS~\cite{zhang2022fast} to accelerate reverse generation. 
However, these approaches face critical limitations: generating high-quality results in fewer than 10 inference steps remains challenging due to the PF-ODE's complex curvature, and discrete approximation errors escalate with increasing data dimensionality, particularly in high-dimensional domains like high-resolution long videos.
Previous methods have explored distillation techniques to enable few-step generation~\cite{salimans2022progressive, meng2023distillation, gu2023boot, yin2024one, song2023consistency}. While these approaches can generate data quickly, they often produce blurry images and suffer from mode coverage issue~\cite{lin2024sdxl}. 
Recent distribution matching techniques~\cite{xu2024ufogen,lin2024sdxl,kong2024act,luo2023diff} introduce auxiliary models to align teacher and student generator distributions, potentially mitigating these problems.
Studies by~\cite{zhou2024score,nguyen2024swiftbrush} have notably demonstrated the potential of one-step image generations that achieve performance surpassing teacher models with multiple steps, and remarkably, they accomplish this in a data-free manner.
However, despite these advances, such distribution matching approaches remain largely unexplored in the context of video diffusion models.

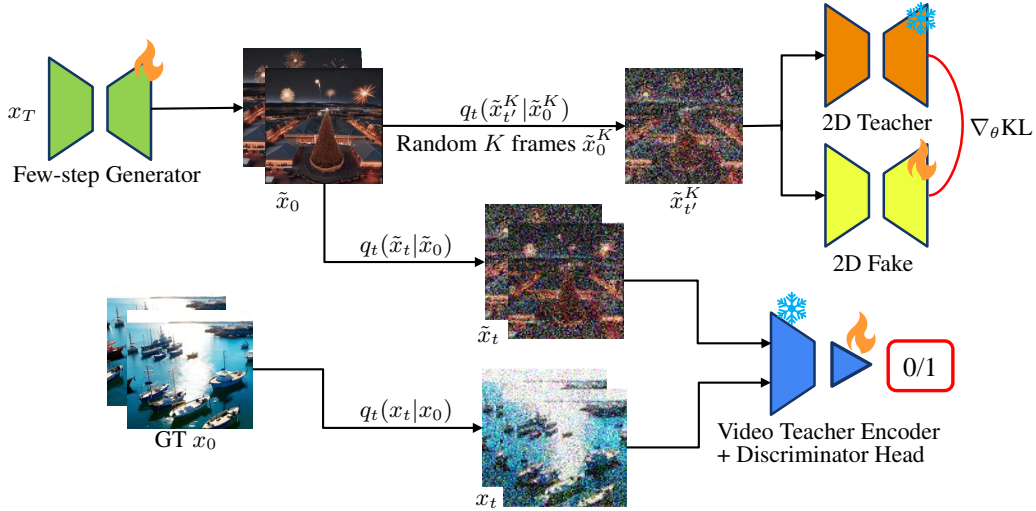
\begin{figure}[!ht]
\centering
\begin{overpic}[width=1.\linewidth]{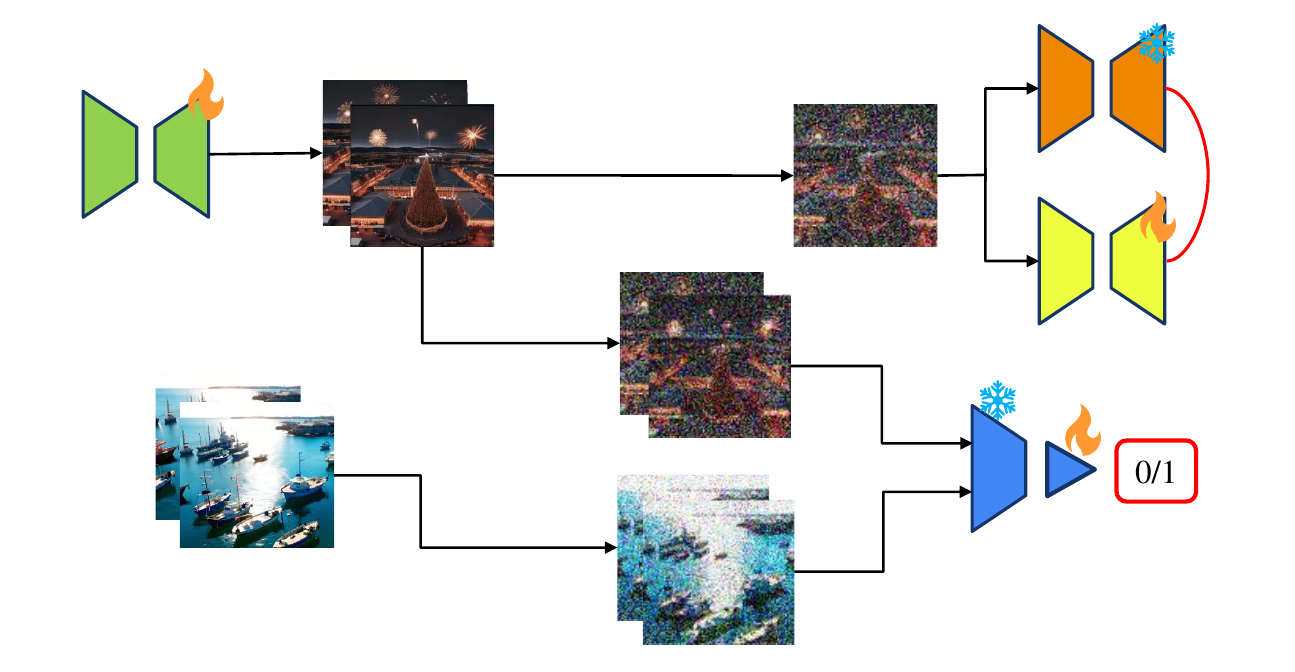}
\put(2.5,39.1){\footnotesize\color{black}{${x}_T$}}
\put(3,33.){\footnotesize\color{black}{Few-step Generator}}
\put(28,30.5){\footnotesize\color{black}{$\Tilde{x}_0$}}
\put(16.5,7.8){\footnotesize\color{black}{GT ${x}_0$}}
\put(36.,10.6){\footnotesize\color{black}{$q_t(x_t|x_0)$}}
\put(36,26.3){\footnotesize\color{black}{$q_t(\Tilde{x}_t|\Tilde{x}_0)$}}
\put(45.5,39.3){\footnotesize\color{black}{$q_t(\Tilde{x}_{t'}^K|\Tilde{x}_0^K)$}}
\put(39.5,36.){\footnotesize\color{black}{Random $K$ frames $\Tilde{x}_0^K$}}
\put(47.2,17.8){\footnotesize\color{black}{$\Tilde{x}_t$}}
\put(47.,2.2){\footnotesize\color{black}{${x}_t$}}
\put(65.5,30.5){\footnotesize\color{black}{$\Tilde{x}_{t'}^K$}}
\put(80,38.){\footnotesize\color{black}{2D Teacher}}
\put(81.,24.5){\footnotesize\color{black}{2D Fake}}
\put(94,37.5){\footnotesize\color{black}{$\nabla_\theta \text{KL}$}}
\put(70,8.5){\footnotesize\color{black}{Video Teacher Encoder}}
\put(70,6.3){\footnotesize\color{black}{+ Discriminator Head}}
\end{overpic}
\caption{
\textbf{Illustration of proposed distribution matching loss}:
The generator produces a video from random noise input. This generated video undergoes forward diffusion to create a noisy video, which is then input to the discriminator for GAN loss computation. Simultaneously, $K$ random frames from the same video are diffused with noise and fed into the 2D teacher and fake model to construct the SDM loss. The discriminator is trained to classify ground truth (GT) videos and generated videos, while the 2D fake model is trained with diffusion loss to learn the generated data's diffusion distribution or score. VAE encoder and decoder are omitted in this figure for simplicity.
}
\label{fig:pipeline}
\vspace{-0.2cm}
\end{figure}

In this work, we propose Accelerate Video Diffusion Model via Distribution Matching (\method{}), a novel framework for distilling knowledge from pre-trained diffusion models and real data into a few-step generator for better single frame quality and motion through distribution matching techniques.
To be specific, initiate from the teacher model, our model is trained with a combination of video GAN loss and 2D \DMD{} loss.
In order to surpass the teacher's performance, we adopt the denoising GAN~\cite{arjovsky2017towards, wang2022diffusion, xu2024ufogen, yin2024improved, lin2024sdxl, zhou2024adversarial} to align the generator's distribution with real data. The discriminator is designed by adding trainable head to the frozen teacher encoder~\cite{wang2022diffusion, yin2024improved, lin2024sdxl, zhang2024sf}.
In addition, we apply 2D \DMD{} loss to ensure single-frame quality, including the proper layout and better prompt-following capability.
This spatial loss approach introduces remarkable flexibility by leveraging the extensive advanced image diffusion models.
Through extensive experiment, we found that our method outperforms the teacher model and previous works in only four sampling steps with AnimateDiff \cite{guo2023animatediff} as teacher text-to-video model.
We illustrate the distillation pipeline of our method in \cref{fig:pipeline}.
\section{Background}
\label{sec:background}

\subsection{Diffusion Models}
Diffusion models define a forward diffusion process that maps data distribution $p_0$ to i.i.d. Gaussian distribution $p_T$ by gradually perturbing the clean data with Gaussian noise.
This process creates intermediate marginal distribution $p_t$ with $t \in (0,T)$.
For data point $x_0 \sim p_0(x_0)$, its noisy version $x_t$ at each time-step $t$ can be drawn from a transition kernel $q_t(x_t|x_0)$:
\begin{equation}\label{eq:forward_diffusion}
    {x}_t = \alpha_t {x}_0 + \sigma_t \epsilon,
\end{equation}
where $\epsilon$ is randomly sampled noise from a standard Gaussian distribution, $\alpha_t$ and $\sigma_t$ are the predefined noise schedule such that $\sigma_0=0$ and $\alpha_T / \sigma_T=0$.

The core learning objective of diffusion models is elegantly simple: training a neural network $\epsilon_\phi$ to reverse the diffusion process, namely, predicting the noise $\epsilon$ that was added to the original data. This is done by minimizing a regression loss:
\begin{equation}\label{eq:dm_objective}
\begin{aligned}
   \mathbb{E}_{{x}_0\sim p_0(x_0), t\sim\mathcal{U}(0,T), \epsilon \sim \mathcal{N}(0,{I})}
   \big[
   \Vert \epsilon_{\phi}(x_{t}, t) - \epsilon \Vert_2^2
   \big],
\end{aligned}
\end{equation}
where $x_t$ is obtained use \cref{eq:forward_diffusion} with clean data ${x}_0$.
While diffusion models $\epsilon_{\phi}$ can be used to remove all the noise with reparameterization $G_\phi(x_{t}, t) = \frac{1}{\alpha_t}(x_t - \sigma_t \epsilon_{\phi}(x_{t}, t))$, we can use the noise-prediction output to move from a highly noisy state $x_t$ to a less noisy state $x_s$ (where $s < t$).
Starting from pure random noise at time $T$, we can repeatedly apply this denoising step to progressively reconstruct a data sample, effectively forming the reverse diffusion process that generates new data.

\subsection{Score Distribution Matching}
In the realm of generative models, distribution matching represents a fundamental paradigm that aims to align the distribution generated by a model with a target distribution of interest.
Formally, the objective is to minimize the Kullback–Leibler (KL) divergence between the generator-induced data distribution $p_{g}$ and the true underlying data distribution $p_{r}$:
\begin{equation}\label{eq:kl}
\begin{aligned}
    {\text{KL}}\left(p_{g} \; \| \; p_{r} \right) &= \mathbb{E}_{x\sim p_g}\left[\log\left(\frac{p_g(x)}{p_r(x)}\right)\right]
    = \mathbb{E}_{\substack{
    \epsilon \sim \mathcal{N}(0; {I}) \\
    x = G_\theta(\epsilon)
    }}\bigg[-\big(\log~p_r(x) - \log~p_g(x)\big)\bigg].
\end{aligned}
\end{equation}
where $G_\theta$ is the generator that transforms data from initial Gaussian noise distribution $p_{\epsilon}$ to the data distribution $p_{g}$.

Score Distribution Matching (SDM), or Variational Score Distillation (VSD) \cite{wang2024prolificdreamer}, is originally proposed to address mode-seeking behavior and over-saturation issue in 3D asserts generation of Score Distillation Sampling (SDS)~\cite{poole2022dreamfusion}.
It is noteworthy that this SDM formulation applies exclusively to the final samples from the generators. This naturally motivates researchers to pursue the development of few-step even one-step generators through a distillation approach, as demonstrated in many recent literature~\cite{luo2023diff, yin2024one, yin2024improved, nguyen2024swiftbrush, dao2025swiftbrush}:
\begin{equation}\label{eq:kl-grad}
\begin{aligned}
\nabla_\theta\mathcal{L}_\text{SDM} &=
    \mathbb{E}_t\left[\nabla_\theta \text{KL}(p_{\text{g},t} \; \| \; p_{\text{r},t}) \right]  \approx \mathbb{E}_{t,\epsilon}\left[w(t)\big(\epsilon_\phi(\Tilde{x}_t, t) - \epsilon_\psi(\Tilde{x}_t, t)\big) \frac{\mathrm{d}G_\theta(\epsilon)}{\mathrm{d}\theta} \right],
\end{aligned}
\end{equation}
where $w(t)$ is a weight function on the loss gradient, $p_{\text{g},t}$ and $p_{\text{r},t}$ represent the noisy marginal probability distributions of the generator and pre-trained model at time step $t$, respectively. $\Tilde{x}_t = \alpha_t G_\theta(\epsilon) + \sigma_t \epsilon'$ describes the noisy state at time step $t$ after forward diffusion with another noise $\epsilon'$. $\epsilon_\phi$ represents the pre-trained noise prediction models and $\epsilon_\psi$ denotes the noise prediction models for the data from generator $G_\theta$ trained with \cref{eq:dm_objective}. The denoising model $\epsilon(x_t,t)$ and the score function $\log p_t(x_t)$ are interconnected through the forward diffusion equation and Tweedie's formula~\cite{robbins1992empirical,efron2011tweedie}.



\section{Related Works}
\label{sec:relatedworks}

\noindent\textbf{Video Diffusion Generative Models.} 
Video Diffusion Generative Models have emerged as a promising approach following the groundbreaking success of diffusion models in image generation. Researchers have explored various strategies to advance video synthesis techniques, addressing challenges in resolution, temporal consistency, and data limitations~\cite{ho2022video, ho2022imagen,blattmann2023stable,zhang2023i2vgen,yu2023video,singer2022make,zhou2022magicvideo,chen2024videocrafter2}.
Multi-stage generation approaches have been used in works like LaVie~\cite{wang2023lavie}, Imagen Videos~\cite{ho2022imagen} and Show-1~\cite{zhang2024show}, typically involving low-resolution initial generation followed by super-resolution refinement. 
VideoCrafter~\cite{chen2023videocrafter1} introduced an innovative approach of treating images as single-frame videos to enhance video quality, while its successor, VideoCrafter2~\cite{chen2024videocrafter2}, proposed a novel data-level disentanglement of motion and appearance to mitigate high-quality video data scarcity.
Temporal modeling has been a critical focus, with approaches like ModelScopeT2V \cite{wang2023modelscope}, VideoLDM \cite{blattmann2023align} and AnimateDiff \cite{guo2023animatediff} implementing temporal motion modules within 2D image models. Notably, AnimateDiff first fine-tunes 2D image models on video frame data, then freezing the UNet and training a separate motion module using the same dataset. 
These models leverage large-scale datasets like WebVid-10M~\cite{bain2021frozen} and Pandas-70M~\cite{chen2024panda} to improve generative capabilities.
The field has seen rapid progression, with recent developments like Open-Sora-Plan~\cite{OpenSoraPlan2024} and Allegro~\cite{zhou2024allegro} pushing the boundaries of video generation by opensourcing models that demonstrates exceptional quality and temporal consistency.


\vspace{0.2cm}
\noindent\textbf{Diffusion Distillation.}
Diffusion distillation techniques aims to reduce the number of function evaluations (NFEs) required to generate samples and can be broadly classified into two primary classes: regression-based~\cite{luhman2021knowledge,song2023consistency,liu2022rectified,salimans2022progressive,gu2023boot,liu2022rectified, meng2023distillation} and distribution-based distillation~\cite{yin2024one,yin2024improved,xu2024ufogen,zhou2024score,zhou2024long,zhou2024adversarial,nguyen2024swiftbrush,dao2025swiftbrush}.
Regression-based methods train the student generator using a regression objective constructed from the PF-ODE, with several distinct approaches emerging in recent research. 
Progressive distillation, as demonstrated by works like \cite{salimans2022progressive,lin2024sdxl}, trains the student to predict directions pointing to the location of teacher's two step prediction and use the student as teacher in the next round.
Consistency models~\cite{song2023consistency,luo2023latent} employ a loss function that constrains student predictions on two consecutive points along the same PF-ODE, ensuring coherent generative behavior from $t=0$ to $t=T$. 
Rectified flow techniques~\cite{liu2022rectified,yan2024perflow,zhu2025slimflow,liu2023instaflow} involve the teacher simulating the entire ODE trajectory or a trajectory segment, with flow matching performed using the endpoints of the interval to get straighter trajectory. 
One the other hand, distribution-based methods model explicitly the score function of the generator or implicitly the density ratio between the generator and pre-trained teacher.
Score distribution matching aims to learn a denoising diffusion model of the few step generator and match it with the teacher diffusion model, as demonstrated in works by~\cite{luo2023diff,yin2024one,zhou2024score}.
Adversarial training emerges as a crucial component in diffusion distillation, serving not only as a primary algorithmic strategy but also as a complementary technique to enhance generation quality~\cite{xu2024ufogen,kim2023consistency,yin2024improved,zhou2024adversarial}.
There is another line of work which extend the idea of SDM based on score identities \citet{franceschi2024unifying, zhou2024score, zhou2024long, zhou2024adversarial, luo2024one}. However, while they demonstrate better performance on certain metrics, it takes much longer time for these methods to converge and sometimes the training is unstable. Moreover, the computation is much slower since the gradient have to backpropogate through the teacher and fake score when update the generator, which is critical for video model distillation as it requires more GPU memory.




\vspace{0.2cm}
\noindent\textbf{Video Diffusion Distillation.}
Recent advancements in video diffusion model acceleration have leveraged diverse distillation techniques to enhance generation efficiency and quality. 
\cite{wang2023videolcm} adopt plain consistency distillation techniques to latent video diffusion models, achieving high-fidelity and temporally smooth video synthesis with four sampling steps.
\cite{wang2024animatelcm} introduce a novel decoupled consistency learning strategy that separates the distillation of image generation priors from motion generation priors, thereby simultaneously improving visual quality and training efficiency.
\cite{zhai2024motion} incorporate motion features into the consistency loss formulation and disentangles motion and appearance learning.
\cite{lin2024animatediff} extend the progressive adversarial diffusion distillation framework proposed in \cite{lin2024sdxl} to distill multiple base models simultaneously, and achieve a better motion module.
\cite{li2024t2v} use mixed reward feedback along with consistency distillation to further improve the quality for few step generation.
For image-to-video generation using Stable Video Diffusion as teacher model, \cite{mao2024osv} utilize LCM and GAN training in two stages to achieve one-step generation, and \cite{zhang2024sf} apply diffusion GAN with an improved discriminator design for state-of-the-art one-step generation.
While our approach shares similarities with \cite{zhang2024sf}, we demonstrate that diffusion GAN alone is not enough for text-to-video distillation, and the 2D SDM loss is crucial for effective model performance.


\section{Methods}

The proposed method aims to distill a few-step video generative model with superior performance compared to the original teacher model's multi-step outputs. 
We introduce a novel pipeline that leverages two complementary distribution matching techniques: Adversarial Distribution Matching (ADM) and SDM. While ADM serves as the primary mechanism, SDM serves as necessary regulation for the distillation process. 
To address the potential mismatch between training and inference when using a few-step generator, we adopt the backward simulation strategy proposed in previous works~\cite{kohler2024imagine,yin2024improved}.

\subsection{Video Adversarial Distribution Matching}
We use ADM to represent GAN-based methods \cite{goodfellow2020generative, arjovsky2017wasserstein}. 
The standard GAN objective is:
\begin{equation}\label{eq:gan_loss}
\begin{aligned}
\min_{G_\theta}\max_{D_{\eta}} \mathbb{E}_{x \sim p_{r}}\left[\log D_{\eta}(x)] + \mathbb{E}_{\epsilon \sim p_{\epsilon}}[\log(1-D_\eta(G_\theta(z)))\right],
\end{aligned}
\end{equation}
where $G_\theta$ is the few-step generator, and $D_\eta$ represents the discriminator with its optimal at $D^\star(x) = \frac{p_r(x)}{p_r(x)+p_g(x)} = \sigma\left(\log\left(\frac{p_r(x)}{p_g(x)}\right)\right)$, which captures the density ratio between the real and generator distributions.
To resolve the issues of non-overlapping probability support between two distributions, we can augment the generator output with a diffusion forward process following the practice in denoising GAN~\cite{wang2022diffusion}:
\begin{equation}\label{eq:denoising_gan_loss}
\begin{aligned}
\min_{G_\theta}\max_{D_{\eta}}&\mathbb{E}_{x \sim p_{r}, t \sim [t_\text{gmin}, t_\text{gmax}]}[\log D_\eta({x}_t,t)]  +  \mathbb{E}_{\epsilon \sim p_{\epsilon}, t \sim [t_\text{gmin}, t_\text{gmax}]}[\log(1-D_\eta(\Tilde{x}_t,t))],
\end{aligned}
\end{equation}
where ${x}_t$ and $\Tilde{x}_t$ are noisy states with noise level corresponding to $t$ and represents the true video data and data generated by generator.
The discriminator $D_\eta$ now takes a noisy video and its noise level $t$ as input, similar to diffusion models.
The noise level $t$ is sampled from the interval $[t_\text{gmin}, t_\text{gmax}]$, where $t_\text{gmin}$ and $t_\text{gmax}$ are hyperparameters. 
The selection of $t_{\max}$ is crucial: it must be sufficiently large to ensure overlapping support between distributions, yet not so large that it becomes challenging for the discriminator to distinguish between real and generated samples.
The loss for generator in the fasion of non-saturation GAN is:
\begin{equation}\label{eq:denoising_gan_g_loss}
\begin{aligned}
\mathcal{L}_{\text{ADM}}(\theta) = \mathbb{E}_{\epsilon \sim p_{\epsilon}, t \sim [t_\text{gmin}, t_\text{gmax}]}\left[\log D_\eta(\alpha_t G_\theta(\epsilon) + \sigma_t \epsilon',t)\right],
\end{aligned}
\end{equation}

In our discriminator architecture, we adopt the UNet encoder from the video teacher model following~\cite{lin2024sdxl}, operating within the latent space to maintain computational efficiency. 
The encoder is kept frozen, while a trainable prediction head generates logits using multi-scale features from the encoder part of the discriminator.
The prediction head extracts features from three spatial scales, processes each through independent convolutional layers, and then concatenates these features. A final convolutional layer transforms the aggregated features into the discriminator's logit output. 
This multi-scale approach enables the discriminator to capture hierarchical representations, improving its capability to distinguish between real and generated video samples.

\subsection{Frame Distribution Matching}




In addition to the video ADM loss, we introduce SDM loss as a second distribution matching loss for distillation. 
We propose using frame-level SDM to regulate individual frame quality, which also enhances the overall distillation efficiency compared to video SDM.
Formally, let a video be represented as $x^{1,2,...,N}$, with $N$ the total number of frames. 
During distillation, we randomly sample $K$ subframes to calculate the SDM loss.
Denoting $\Tilde{x}^K_t = \Tilde{x}^{k_1,k_2,..,K}_t$ and $G^K_\theta(\epsilon)=G_\theta(\epsilon)^{k_1,k_2,..,K}$, the gradient of the frame SDM can be written as:
\begin{equation}\label{eq:frame_SDM}
\begin{aligned}
\nabla_\theta\mathcal{L}_\text{SDM} = &\mathbb{E}_{t,\epsilon}\bigg[w(t)\big(\epsilon_\phi(\Tilde{x}^K_t, t) - \epsilon_\psi(\Tilde{x}^K_t, t)\big) \frac{\mathrm{d}G^K_\theta(\epsilon)}{\mathrm{d}\theta} \bigg].
\end{aligned}
\end{equation}
In our approach, the SDM loss is applied only to the final predicted clean samples from the generator. 
This allows us to leverage any existing 2D diffusion model with shared latent space to take the noisy input samples and construct the SDM loss.

\vspace{0.2cm}
\noindent\textbf{The overall training objective.}
The student generator $\mathbf{v}_\theta$ is trained to minimize the combination of the aforementioned two distribution matching loss terms:
\begin{equation}\label{eq:distill_loss}
\begin{aligned}
   \mathcal{L}(\theta):= \lambda_{\text{SDM}} \mathcal{L}_{\text{SDM}}(\theta) + \lambda_{\text{ADM}}\mathcal{L}_{\text{ADM}}(\theta),
\end{aligned}
\end{equation}
where $\lambda_{\text{SDM}}$ and $\lambda_{\text{ADM}}$ are the importance weight for SDM loss and ADM loss, respectively.
The discriminator $D_\eta$ is trained to optimize \cref{eq:denoising_gan_loss} and the SDM fake model $\epsilon_\phi$ is trained with \cref{eq:dm_objective}.
To train the generator more stably, we employ a common two-timescale update rule from GAN training. Specifically, we train the discriminator and 2D fake model twice as often as the generator. 
The distillation algorithm of the proposed method is summarized in \cref{alg:distill}.

\begin{figure*}[t!]
\centering
\begin{overpic}[width=1.\linewidth]{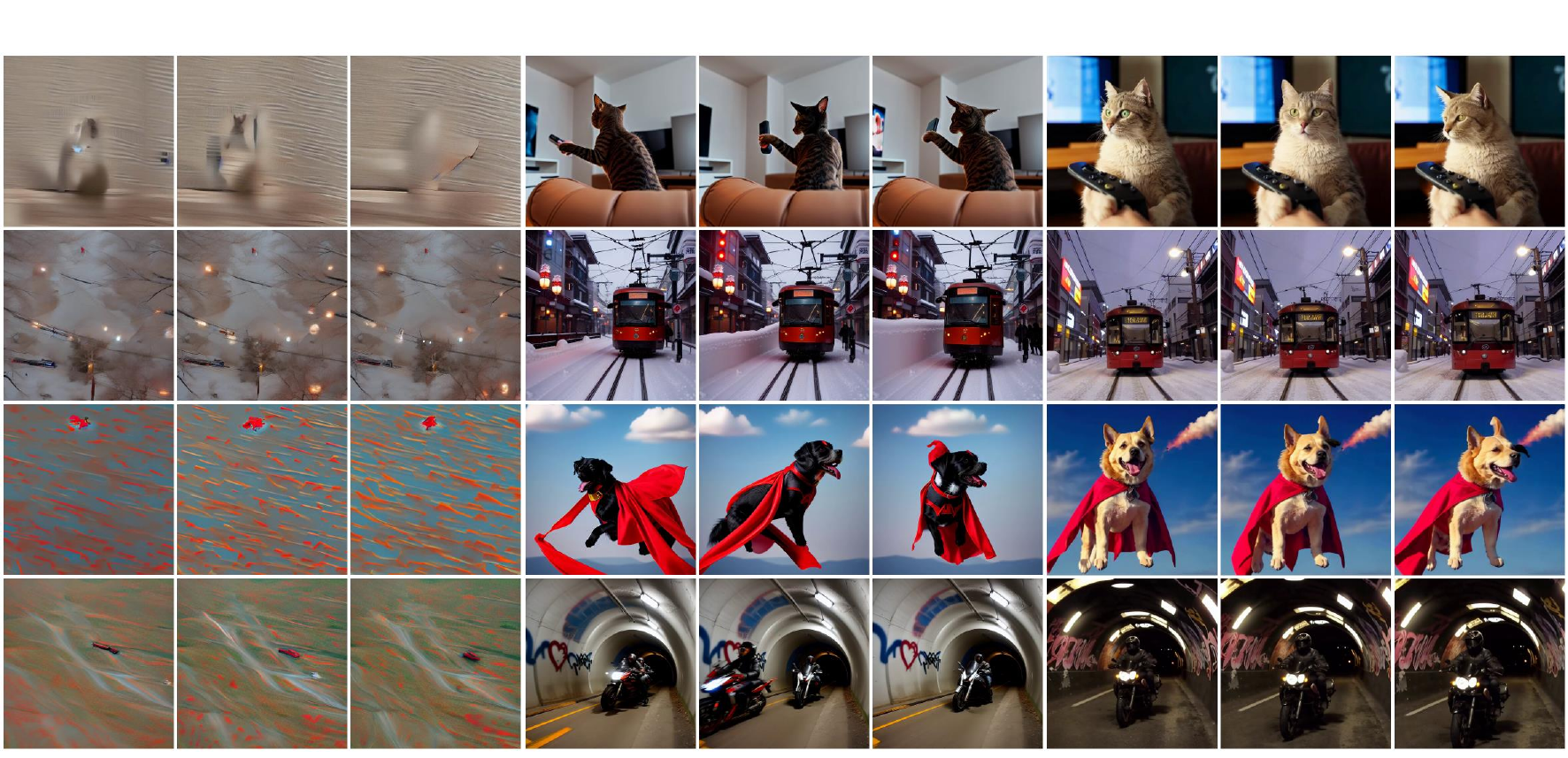}
\put(4.4,45.5){\color{black}{\small Teacher 4 steps (CFG=7.5)}}
\put(37.9,45.5){\color{black}{\small Teacher 25 steps (CFG=7.5)}}
\put(73.5,45.5){\color{black}{\small Our 4 steps (no CFG)}}
\end{overpic}
\vspace{-0.25cm}
\caption{ 
Comparison between our method and teacher AnimateDiff model with different sampling steps.
We display the 1st, 8th and last frame.
}
\label{fig:comparison_teacher}
\vspace{-0.2cm}
\end{figure*}

\begin{algorithm}[H]
    \small
   \caption{\method{} Distillation}
   \label{alg:distill}
    \begin{algorithmic}[1]
     \Require video teacher model $\epsilon_{\phi}$, 2D teacher model $\epsilon_{\phi2D}$, dataset $\mathcal{D}$, forward diffusion process $p(x_t|x_0)$, 
     $t_\text{min}$, $t_\text{max}$, $t_\text{gmin}$, $t_\text{gmax}$,
     $\lambda_{\text{SDM}}$, $\lambda_{\text{AMD}}$, sub-frame number $K$, $w(t)$
    \State{Initialize the student generator $G_{\theta}$ with the weights of $\epsilon_{\phi}$, discriminator with the encoder of $\epsilon_{\phi}$}; initialize 2D fake model $\epsilon_{\psi_{2D}}$ with $\epsilon_{\phi_{2D}}$
    \Repeat
        \State{sample $z_T \sim \mathcal{N}(\mathbf{0},\mathbf{I})$}
        \State{Run backward simulation to get $G_\theta(z_T)$}
        \State{Randomly sample new noise $\epsilon \sim \mathcal{N}(\mathbf{0},\mathbf{I})$, $t \sim (t_\text{gmin}, t_\text{gmax})$}
        \State{Calculate $\Tilde{x}_{t}$ using forward diffusion}
        \State{Calculate ADM loss using \cref{eq:denoising_gan_g_loss} }
        \State{Randomly sample new noise $\epsilon \sim \mathcal{N}(\mathbf{0},\mathbf{I})$, $t' \sim (t_\text{min}, t_\text{max})$, random $K$ frames}
        \State{Calculate $\Tilde{x}^K_{t'}$ using forward diffusion}
        \State{Calculate SDM loss using \cref{eq:frame_SDM}}
        \State{Update $\theta$ with gradient-based optimizer using $\nabla_\theta\left(\lambda_{\text{SDM}} \mathcal{L}_{\text{SDM}}(\theta) + \lambda_{\text{ADM}}\mathcal{L}_{\text{ADM}}(\theta)\right)$.}
        \State{Randomly sample new noise $\epsilon \sim \mathcal{N}(\mathbf{0},\mathbf{I})$, $t \sim (t_\text{min}, t_\text{max})$}
        \State{Calculate $\Tilde{x}_{t}$ using forward diffusion}
        \State{Update ${\phi_{2D}}$ by optimizing the diffusion loss \cref{eq:dm_objective}}
        \State{Randomly sample new noise $\epsilon \sim \mathcal{N}(\mathbf{0},\mathbf{I})$, $t' \sim (t_\text{gmin}, t_\text{gmax})$}
        \State{Calculate $\Tilde{x}_{t'}$ using forward diffusion}
        \State{Update ${\eta}$ by maximizing the GAN objective \cref{eq:gan_loss}}
    \Until{\textit{convergence}}
    \State{\textbf{Return} few-step generator $G_{\theta}$}
    \end{algorithmic}
\end{algorithm}

\section{Results and Discussion}
\label{sec:experiments}

\begin{figure*}[t!]
\centering
\begin{overpic}[width=0.98\linewidth]{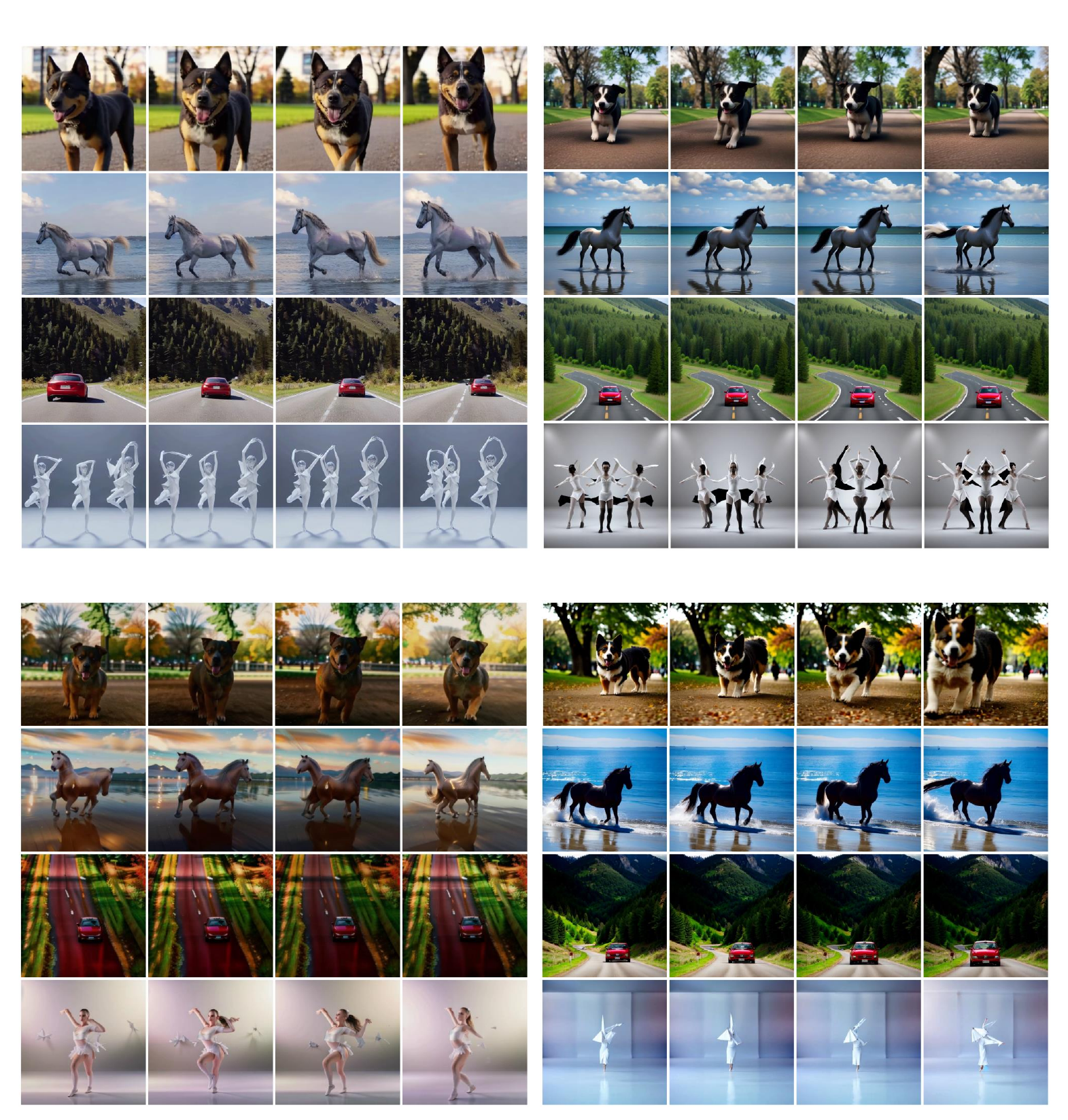}
\put(14.7,97.){\color{black}{\small Our 4 steps generation}}
\put(52.2,97.){\color{black}{\small AnimateDiff-Lightning 4 steps generation}}
\put(4.4,47.2){\color{black}{\small Motion Consistency Model 4 steps generation}}
\put(55.9,47.2){\color{black}{\small AnimateLCM 4 steps generation}}
\end{overpic}
\vspace{-0.2cm}
\caption{\textbf{Qualitative comparison on base model AnimateDiff.} 
From top to bottom the text prompts are: 
1) a dog with big expressive eyes running in a city park;
2) a majestic horse with a long flowing tail running at a tranquil beach;
3) a red car, moving on the road, mountain, green grass and trees;
4) Origami dancers in white paper, 3D render, ultra-detailed, on white background, studio shot, dancing modern dance.
}
\label{fig:comparison_main}
\vspace{-0.2cm}
\end{figure*}
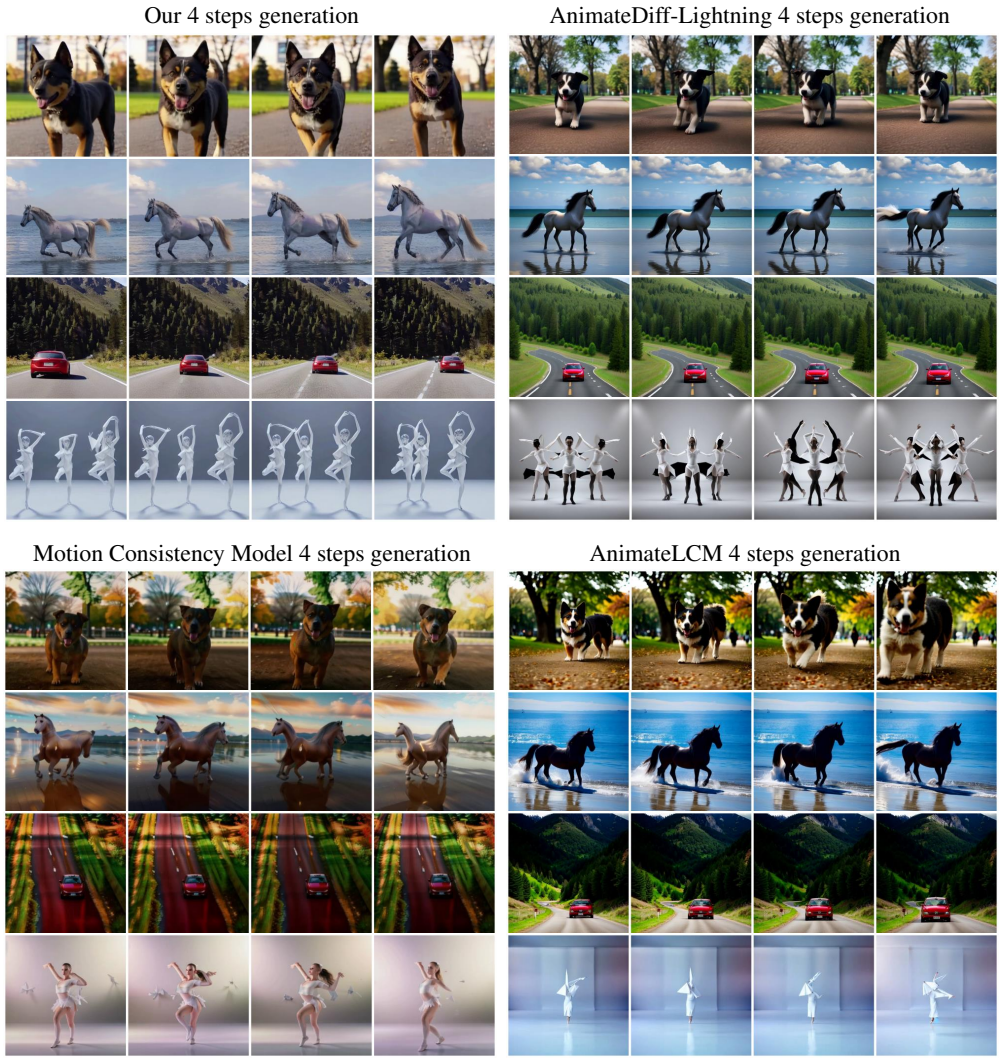

In this section, we provide experimental details and empirical results of \method{} and compare it with prior arts. 

\subsection{Experimental Setup}
\noindent\textbf{Datasets.}
We use real video dataset for ADM training.
We use the open sourced dataset OpenVid-1M~\cite{nan2024openvid}, which contains $\sim$1M high quality videos, without any filtering.
We also optioned our internal dataset at the same scale and achieved similar performance.
For validation, we randomly select 100 videos and the corresponding text prompts from the WebVid10M validation set~\cite{bain2021frozen} to evaluate the metrics.

\vspace{0.2cm}
\noindent\textbf{Compared Methods.}
We mainly compare our method with previous state-of-the-art distilled video generation models and teacher models.
For the AnimateDiff teacher \cite{guo2023animatediff}, we compare with AnimateLCM~\cite{wang2024animatelcm}, AnimateDiff-Lightning~\cite{lin2024animatediff}, and Motion Consistency Model~\cite{zhai2024motion}.
Our comparative evaluation employs both qualitative visual assessment and quantitative metrics, specifically the Fréchet Video Distance (FVD)~\cite{unterthiner2018towards} and CLIPScore~\cite{hessel2021clipscore}, applied to the validation dataset.
For the FVD calculation, we first preprocess the validation data by resizing, center cropping, and subsampling 16 frames to get video of same shape as the model output, then compare the distribution characteristics between the validation and model-generated video samples.
The CLIPScore evaluates the semantic alignment and visual quality of the generated content, where we compute the average score for each frame within a video with its text prompt and then report the mean CLIPScore across all generated videos. 
This method offers insights into the semantic alignment and visual quality of the generated content.

\vspace{0.2cm}
\noindent\textbf{Training Details.}
We initialize the generator with the video teacher model with the same architecture, and initialize the 2D fake score with the 2D teacher model with the same architecture.
By default we use AnimateDiff with Realistic Vision as teachers.
For training stability, the output of discriminator is the logits.
We use the AdamW optimizer with a linear warm-up schedule over 500 training steps, followed by a learning rate of 2e-5 for the motion module and 4e-6 for the 2D UNet backbone to train our model. 
Adamw optimizer is also used for the discriminator and fake score, with a learning rate of 2e-5.
Betas of [0.9,0999] and weight decay of 0.01 are used for all optimizers.
No EMA is used during the experiments for training efficiency.
We use $t\sim \mathcal{U}[0.2T,0.98T]$ in SDM loss and $t\sim \mathcal{U}[0,0.5T]$ in ADM loss.
We sample 16 frames with a temporal stride of 4 and crop a 512 × 512 center region after resize from each source video as model input. 
The entire network is trained on 8 NVIDIA H800 GPUs with a batch size of 2 on each card, requiring approximately 5k training iterations to produce relatively reliable video generation results and we use the checkpoint from 10k iteration for evaluation.

\begin{figure*}[t!]
\centering
\begin{overpic}[width=1.\linewidth]{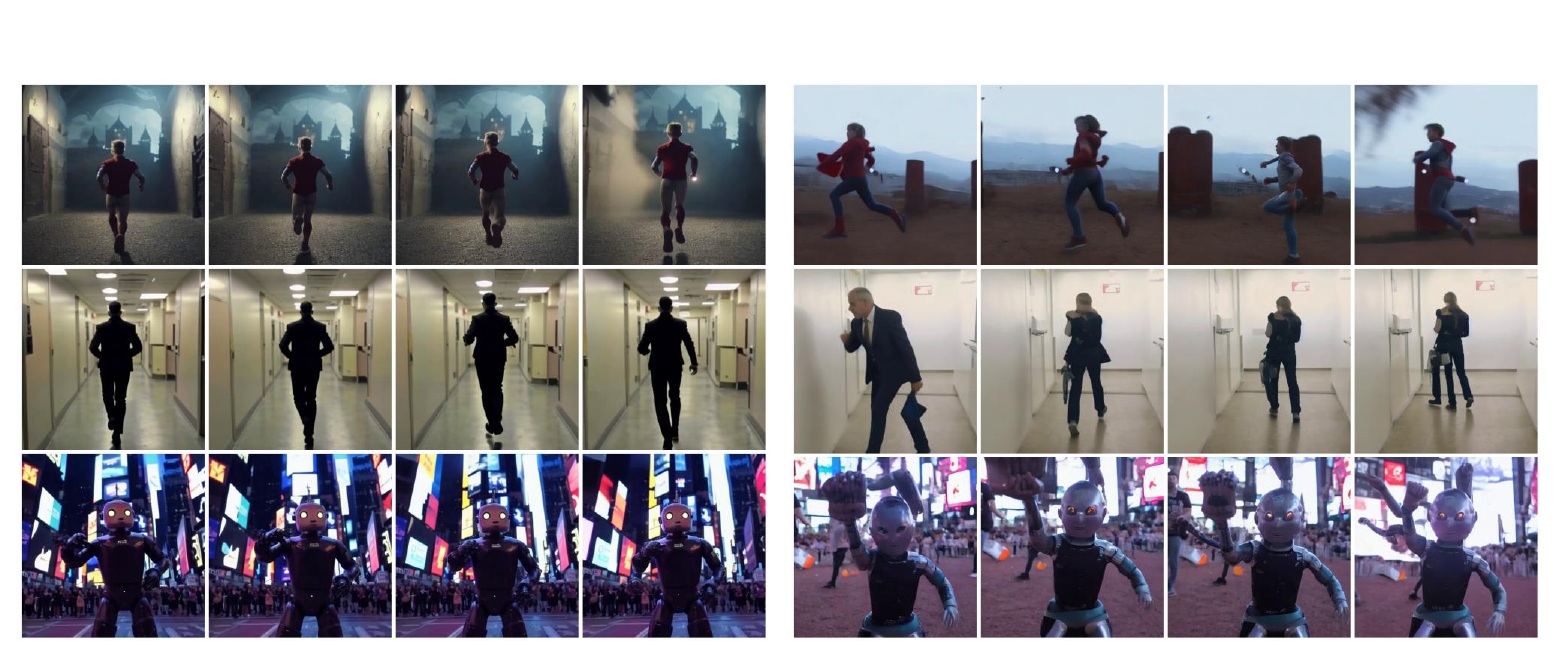}
\put(23.4,37.3){\color{black}{\small Our}}
\put(64.9,37.4){\color{black}{\small Diffusion GAN Alone}}
\end{overpic}
\vspace{-0.4cm}
\caption{ 
Comparison between our method and diffusion GAN alone training on 4 step generation.
}
\label{fig:comparison_gan_only}
\end{figure*}

\begin{table}[t!]
\centering
\footnotesize
\resizebox{0.99\linewidth}{!}{
\begin{tabular}{lccc}
\toprule
 Method & NFEs ($\downarrow$) & FVD ($\downarrow$)  & CLIPScore ($\uparrow$) \\
\midrule 
 Motion Consistency Model \cite{zhai2024motion} & 4 & 1765.30 &  28.60 \\
 AnimateLCM \cite{wang2024animatelcm} & 4  & 1405.79 &  28.44  \\
 AnimateDiff-Lightning \cite{lin2024animatediff} & 4 & 1623.98 & 29.47  \\
 Ours (with video SDM) & 4 & {1273.13} & {30.56}  \\
 Ours (with 2D SDM) & 4 & \textbf{1271.45} & \textbf{32.01}  \\

\bottomrule
\end{tabular}}
\caption{
\label{tab:metric}
Metric comparison of different methods with 4 sampling steps.
}
\vspace{-0.2cm}
\end{table}

\subsection{Main Results}

\noindent\textbf{Comparison with Teacher.}
In \cref{fig:comparison_teacher}, we present comparative results for video generation between our proposed method and the baseline AnimateDiff approach. The results demonstrate our method's remarkable ability to generate high-quality videos using just 4 NFEs, whereas the baseline method requires 25 DDIM sampling steps to achieve comparable results.
Qualitative analysis reveals that our approach generates more realistic videos with reduced temporal distortion, as evidenced in the first and third rows of the comparison figure.
Furthermore, our distilled models do not need the Classifier-Free Guidance (CFG) during inference, further improving the sampling efficiency.

\begin{figure*}[t!]
\centering
\begin{overpic}[width=1.\linewidth]{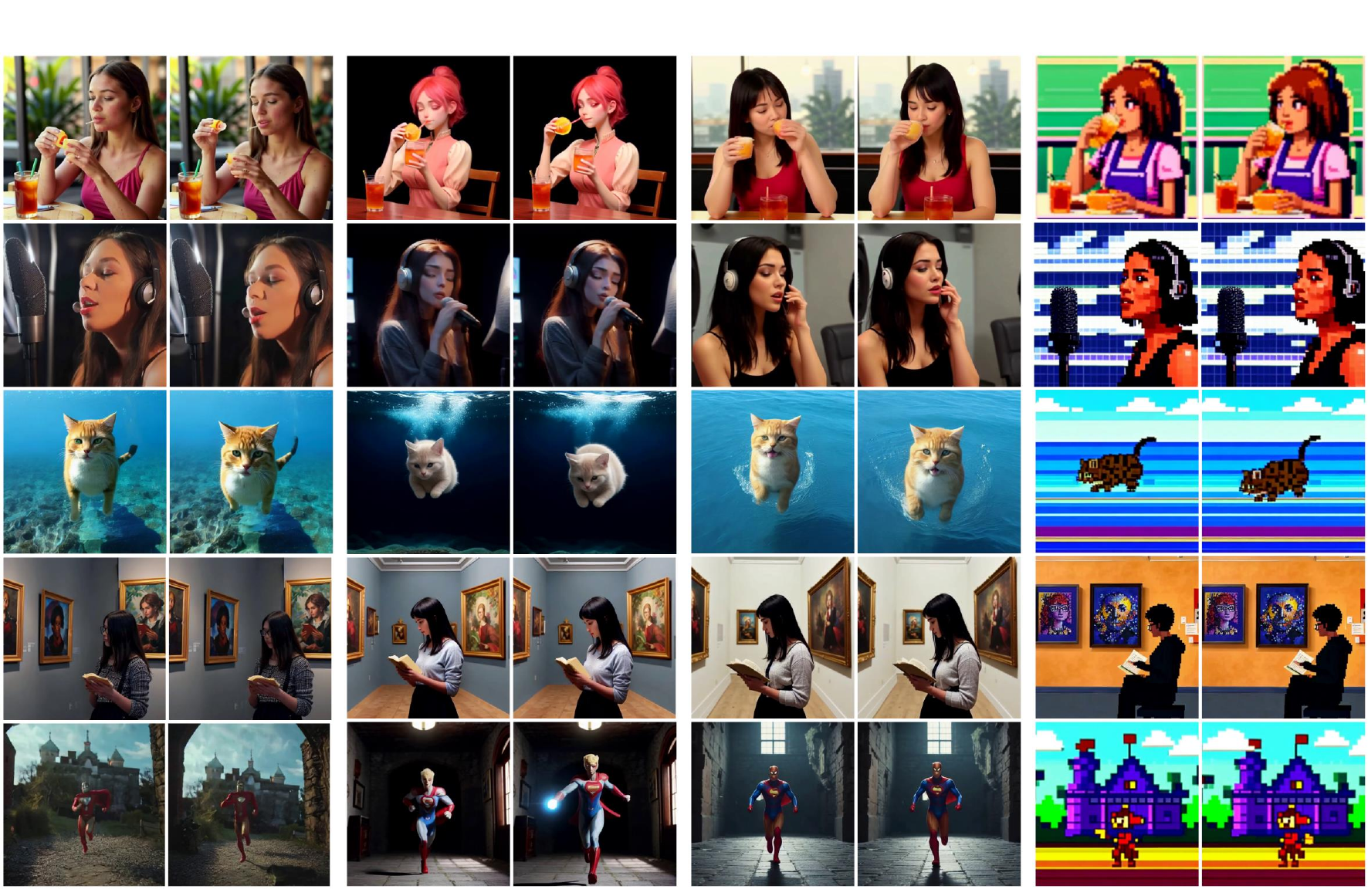}
\put(4.8,62.2){\color{black}{\small Realistic Vision}}
\put(33.4,62.2){\color{black}{\small Toonyou}}
\put(56.2,62.2){\color{black}{\small Dreamshaper}}
\put(83.9,62.2){\color{black}{\small Pixel-art}}
\end{overpic}
\vspace{-0.4cm}
\caption{ 
Visual results of our method with different 2D SDM models.
}
\label{fig:comparison_2D}
\vspace{-0.2cm}
\end{figure*}
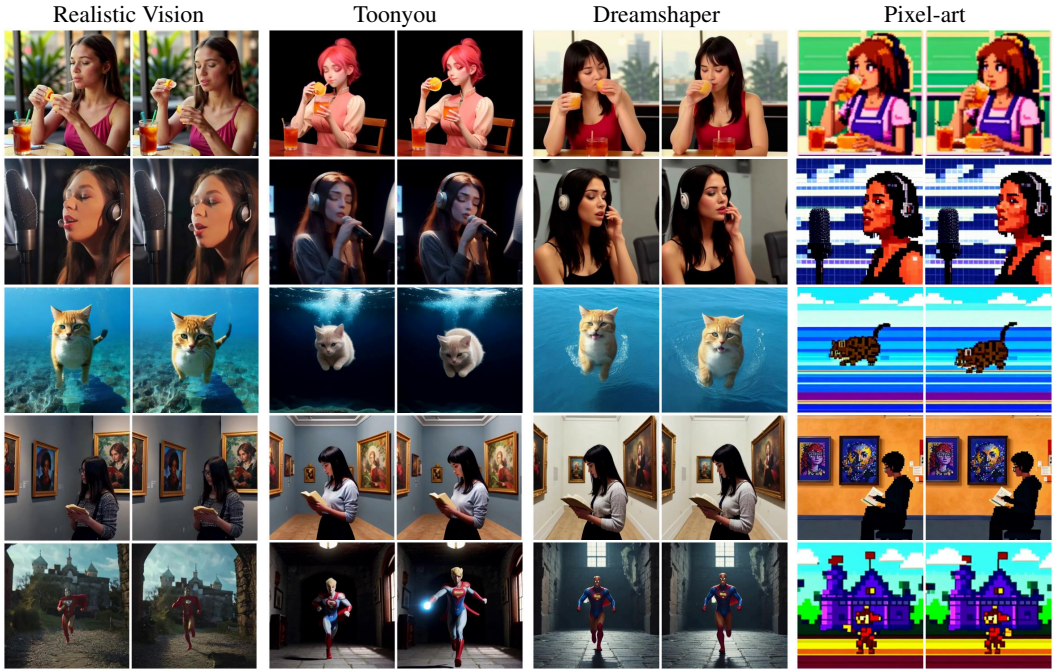

\vspace{0.2cm}
\noindent\textbf{Comparison with Benchmarks.}
\cref{tab:metric} and \cref{fig:comparison_main} present the visual results and metrics on the validation dataset.
The comprehensive evaluation reveals our approach's remarkable capability to generate high-fidelity and smooth videos.
As demonstrated in \cref{tab:metric}, our model significantly outperforms competing methods, achieving substantially better scores in both FVD and CLIPScore. 
In \cref{fig:comparison_main}, we present a side-by-side comparison of evenly sampled frames generated using the same text prompts across different methods. 
Notably, the competing methods exhibit noticeable flickering effects that our evaluation metrics fail to capture.
In contrast, our approach delivers videos characterized by exceptional temporal consistency and photorealistic visual clarity.
This discrepancy highlights the limitations of existing quantitative assessment techniques and emphasizes the importance of comprehensive, multi-dimensional evaluation in video generation research.
Besides, we conduct a metric comparison with both 2D SDM loss and video SDM loss, as presented in \cref{tab:metric}, and reveal remarkably consistent FVD and better CLIPScore, which suggests the robustness of our approach and indicates the effectiveness os proposed method. 






\vspace{0.2cm}
\noindent\textbf{Importance of SDM.}
In \cref{fig:comparison_gan_only}, we visually demonstrate the necessity of the 2D SDM loss in our method. When the model is distilled using only the diffusion GAN loss, we observe significant temporal inconsistencies. These inconsistencies manifest as severe distortions, identity changes.
We also notice a style shift from the teacher to the data distribution, this can be mitigated by the 2D image model when its style is closer to the video teacher model.

\vspace{0.2cm}
\noindent\textbf{Result with Different 2D model.}
The advantages of employing 2D \DMD{} instead of video \DMD{} for video model distillation extend beyond computational efficiency. 
As illustrated in \cref{fig:comparison_2D}, this approach offers significant flexibility to generate video with different styles, primarily due to the vast ecosystem of existing 2D diffusion models that can be readily leveraged.
This strategy opens up new avenues for cross-domain model adaptation and knowledge transfer.
The visual results demonstrates that the 2D \DMD{} approach presents a versatile solution to video model distillation, bridging the gap between computational constraints and generative performance.






\section{Limitations and Future Works}
\label{sec:limitaion}


While our proposed \method{} achieves high-quality video generation using as few as 4 sampling steps, we acknowledge several limitations that present opportunities for future research.
First, we found it challenging to distill a one-step generator, which tends to exhibit a noticeable flickering effect, similar limitation is observed in other methods on the same teacher model.
Additionally, our approach requires maintaining two auxiliary models during the training process, which can reduce overall training efficiency and poses challenges for scaling to larger and more complex models.
Furthermore, we observed a degradation in the diversity of the distilled generator's output compared to the original teacher model.
These challenges represent promising avenues for future investigation, and we intend to explore them in our upcoming work.


\section{Conclusion}
\label{sec:conclusion}


In this work, we propose a novel method, denoted as \method{}, which leverages two distribution matching losses to enhance the quality of generated videos while accelerating the video generation process. Specifically, we utilize adversarial distribution matching to enable the model to produce high-quality results with few sampling steps. Additionally, we employ score distribution matching to regulate the shape and structure of individual frames, further improving the overall video quality.
Our experimental results demonstrate that our distilled model is able to produce videos with superior frame quality compared to the teacher models, while requiring only 4 sampling steps during inference. 
This work demonstrates the power of distribution matching methods on distillation of video diffusion models, contributes to the growing community of research exploring efficient and high-quality video generation.


\noindent\textbf{ACKNOWLEDGEMENTS.}
We thank Tianwei Yin for his insightful discussion during this work.


\bibliography{references}


\end{document}